\documentclass[letterpaper, 10 pt, conference]{ieeeconf}  
\IEEEoverridecommandlockouts                              

\usepackage[utf8]{inputenc}
\usepackage[normalem]{ulem}
\usepackage{cite}
\usepackage{graphicx}
\usepackage{caption}
\usepackage{subcaption}
\usepackage{xcolor}
\usepackage{url}
\usepackage{amssymb}
\usepackage{amsmath}
\usepackage{algorithm}
\usepackage{algorithmic}
\useunder{\uline}{\ul}{}
\usepackage{multirow}
\usepackage{float}

\newcommand{\taskloss}{\textit{task loss }}
\newcommand{\jointloss}{\textit{joint loss }}

\title{\LARGE \bf
Leveraging Forward Model Prediction Error for Learning Control
}

\author{Sarah Bechtle$^{1}$ Bilal Hammoud$^{1,2}$ Akshara Rai$^{3}$ Franziska Meier$^{3}$ and Ludovic Righetti$^{1,2}$
\thanks{$^{1}$Max Planck Institute for Intelligent Systems, Tübingen, Germany
        {\tt\small sbechtle@tuebingen.mpg.de}}%
\thanks{$^{2}$Tandon School of Engineering, New York University, Brooklyn, NY}%
\thanks{$^{3}$Facebook AI Research, Menlo Park, CA}%
\thanks{This work was in part supported by the European Union's Horizon 2020 research, innovation program (grant agreement 780684 and European Research Councils grant 637935), the National Science Foundation (grants 1825993, 1932187 and 2026479). Sarah Bechtle was in part supported by the International Max Planck Research School for Intelligent Systems.}
}

\begin{document}

\maketitle
\thispagestyle{empty}
\pagestyle{empty}

\begin{abstract}
Learning for model based control can be sample-efficient and generalize well, however successfully learning models and controllers that represent the problem at hand can be challenging for complex tasks. Using inaccurate models for learning can lead to sub-optimal solutions, that are unlikely to perform well in practice. In this work, we present a learning approach which iterates between model learning and data collection and leverages forward model prediction error for learning control. We show how using the controller's prediction as input to a forward model can create a differentiable connection between the controller and the model, allowing us to formulate a loss in the state space. This lets us include forward model prediction error during controller learning and we show that this creates a loss objective that significantly improves learning on different motor control tasks. We provide empirical and theoretical results that show the benefits of our method and present evaluations in simulation for learning control on a 7 DoF manipulator and an underactuated 12 DoF quadruped. We show that our approach successfully learns controllers for challenging motor control tasks involving contact switching.
\end{abstract}

\section{INTRODUCTION}
Data driven model based approaches to learn control have been introduced as an alternative to overcome the limitations of imperfect analytical models of the robotic tasks \cite{bristow2006survey}. In this work, we consider iterative model based learning, where we iterate between learning a forward dynamics model of the robot, and using it to learn a controller. The controller essentially inverts the forward model, computing an action given a current and desired next state. Two challenges arise here: first the quality of the learned forward model is decisive for the success of learning control and second, it might be challenging to learn a controller based on the learned model, in order to perform a task successfully. We present an approach that couples model and controller learning by leveraging forward model prediction error during controller learning. The controller predicts the motor command required to achieve a desired state. The forward model predicts the next state, from the current measured state and motor command predicted by the controller, thus representing the causal relationship of the movement \cite{wolpert1995internal}. This connects the models, as the predicted action is used as input to the forward model.
From a robotics perspective, forward and inverse models are representations of the physical properties of the robot. In the neuroscience and cognitive science literature \cite{miall1996forward, ito1970neurophysiological}, the presence of internal models, as representation of the body in the human brain \cite{ishikawa2016cerebro} are believed to play an important role.
In \cite{wolpert1998internal} the authors explain the necessity for a connection between forward and inverse models in the cerebellum by pointing out that acquiring an inverse model purely from motor learning is difficult, since the optimal motor command is not available during learning (otherwise the learning would not be necessary). From a robotics perspective, this argument holds as well, since a desired trajectory is usually defined in the state space and not in the action space.

Following this observation, and unlike the more common approach in the robotics literature, where the forward or the inverse model are trained separately using supervised learning from data (as for example in \cite{nguyen2008computed,camoriano2016incremental,atkeson1988using,miller1987sensor}), we show how connecting the models, and formulating a loss in the state space, improves the performance when learning control. In contrast to other work \cite{jordan1992forward} that considers learning these models together, we show how including the prediction error of the forward model during controller learning creates an unbiased loss signal, that leads to a significant improvement in performance.

In a nutshell, the contributions of this paper are: 1) We explore the effects of connecting controller and forward model during learning control in an iterative fashion on a manipulator and a quadruped. 2) We show, with theoretical and empirical results, how including forward model prediction error during controller learning significantly improves learning a motor control task on a robot. We present manipulation and locomotion experiments, specifically we also show learning of a walking controller that can inherently handle contact switching. 
\begin{figure}[h]
    \centering
    \includegraphics[width=0.4\textwidth]{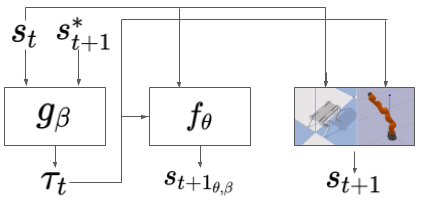}
    \small
    \caption{Overview of connecting inverse and forward models: the motor command, that is the output of the controller, is fed to the forward model used to predict the next state. The motor command is also run on the robot in order to observer the real next state. The learned controller is then updated by taking the gradient of either \eqref{eq:task_loss} or \eqref{eq:joint_loss}}
    \label{fig:coupled_models_overview}
\end{figure}{}

\section{Related Work}
\subsection{Coupling forward and inverse models}
Wolpert et al. present in \cite{wolpert1998multiple} an architecture for multiple paired inverse and forward models, the pairs are coupled and trained jointly. The predictions of the forward models determine which inverse model to use. \cite{koert2018learning} extends this for a manipulator. \cite{schillaci2012coupled} present results on coupled learning of kinematic models for tool use.  In \cite{lutter2019deep} the authors present a deep neural network that structures the learning of a manipulator's dynamics model following Lagrangian mechanics. The trained model can be used for forward as well as inverse dynamics computation, but does not directly connect the models.
Most similar to our work is \cite{jordan1992forward}, where the authors show the benefits of using a `distal teacher' for training the inverse model on a 2 link 2D arm. Their approach is based on a stochastic gradient, computed by comparing the observed states with the desired state. In contrast to these approaches, we present an iterative method to train the models jointly. Our experiments are conducted on two different robots, in 3D, and present a loss function that considers the forward model prediction error during controller learning. We show in Section \ref{sec:approach} how our approach mathematically differs from \cite{jordan1992forward}, and in Section \ref{sec:experiments} that it achieves significantly better results on higher dimensional systems. In particular, our approach can easily include contact interactions.
\subsection{Using model prediction error for learning}
The idea of using model prediction error during learning has been explored within the reinforcement learning literature mostly form the perspective of intrinsically motivated agents. For example, \cite{Barto2004, Singh2004, Singh2010} propose rewarding agents to minimize prediction errors of sensory events to explore the state space. This work is limited to low-dimensional and discrete state-and-action spaces. More recently \cite{Bellemare2016, Pathak2017,tanneberg2019intrinsic, laversanne2018curiosity} present results on higher dimensional systems, however this work focuses on model free reinforcement learning where the learned models are purely used to provide an additional learning signal to train a policy. In contrast to this work, our approach uses forward model prediction error during learning in a setting where the learned model is actually used to learn a motor control task. 
\subsection{Improving model learning in model based approaches}
Fewer works have included additional learning signals during model based learning. \cite{shyammax} proposes a measure of disagreement in an ensemble of forward models as an exploration signal. \cite{bechtle2020curious} shows that including the predictive uncertainty of the forward model during controller optimization could improve forward model learning.
In \cite{lopes2012exploration}, an empirical measure of learning progress in included on a low dimensional discrete MDP. Similarly, self correcting forward models were proposed in \cite{talvitie2014model,talvitie2016self} but the considered problem remains low dimensional. 
While it is widely acknowledged that model quality is of crucial importance in model based approaches, to the best of our knowledge this problem is seldom tackled for high dimensional systems.
\subsection{Learning models including force measurements}
Learning models that include non-trivial contact interactions is especially challenging as contacts create discontinuous force measurements and control actions.
In \cite{zhang2019leveraging} the authors use force measurements as an additional input to their model for a manipulator. However, the measurements are not used for controller learning but only to discriminate between different tasks. In \cite{lee2019making} multimodal input signals, including forces, are used to train an embedding for a downstream model free reinforcement learning task that takes as input the learned embedding but does not use the learned model during policy learning. 
Even with accurate physical models, the conception of inverse dynamics controllers is challenging with changing contacts \cite{herzog2016momentum} as special care is necessary at each contact transitions, i.e. typically involving manual design of switching events or advanced constraint switching strategies \cite{jarquin_task_hierarchy2013}.
We show in Section \ref{sec:experiments} how our approach enables to learn a walking controller for a quadruped by including measured contact forces not only as inputs, but also as predictions during controller learning. Importantly, the learned controller seamlessly handles contact switches without any additional assumptions as it learns to predict contact switches using the forward model.
\section{Problem Formulation and Approach}\label{sec:approach}
The goal of model based learning control is to learn a forward model $f$ of the dynamics of the robot and a controller, or inverse model, $g$. In general, $g$ can be learned from data but can also be optimized using trajectory optimization algorithms see \cite{levine2013guided,deisenroth2011pilco,bechtle2020curious} for a variety of approaches of iteratively learning a model and a controller.

In this work, we propose an algorithm inspired by the concept of connected forward and inverse models, while still being able to iteratively collect data and update the models. We learn a forward model $f_\theta$ that performs one step prediction of the form $s_{t+1} = f_{\theta}(s_t,\tau_t)$, where $\theta$ are the parameters of the forward model, $s_t$ and $\tau_t$ the state and action at time $t$. We also learn a controller $g_\beta$ that predicts $\tau_t = g_{\beta}(s_t,s^*_{t+1})$, given the current state $s_t$ and the desired state $s^*_{t+1}$. $\beta$ are the parameters of the controller and $s^*$ can be the immediate desired next state, or a final goal state. 
We learn both models from data collected on the robot,while alternating between model learning and data collection. Algorithm \ref{algo:mbrl_loop_coupled} shows the training procedure. We create a direct connection between $f_\theta$ and $g_\beta$ by using the action predicted by $g$ as an input to $f$. Since $s_{t+1}{_{\theta,\beta}} = f_\theta(s_t,g_{\beta}(s_t,s^*_{t+1}))$, the next state is a function not only of the parameters of $f$ but also of $g$.
This means that, using $s_{t+1}{_{\theta,\beta}}$ we can formulate a loss that enables us to compute a gradient to update the parameters $\beta$ of $g$. 

In Figure \ref{fig:coupled_models_overview} the coupling of the forward and the inverse model is illustrated. Using $s_{t+1}{_{\theta,\beta}}$ has the advantage of representing the actual effect that the action, that was predicted by $g$ has. In contrast to learning $g$ in a supervised fashion from collected data, this approach is conceptually more sound as the correct or desired supervision signal for the action is usually not available. However the goal of the task, $s^*_{t+1}$, is available in the state space. 

In model based approaches, forward models and controllers are inherently intertwined: during the training phase, the forward model predicts the possible next state, and the controller is learned based on this prediction. The controller is the acting component of the loop, facilitating data collection on the robot that is used to update the models.
It becomes clear here, that if the forward model prediction is inaccurate, controller training will fail and the collected data might not be meaningful for the current task. This brings us back to one of the major challenges of model based learning, which is to learn models that are accurate enough to use to act on a robot. 

In the next section, we introduce a new loss function as well as other, more standard, losses used as comparison. We propose a loss function for controller learning that ultimately reduces model bias, by including forward model prediction error for learning control. As a result, this improves model prediction and task performance.
\subsection{Learning control via coupled models with \jointloss}\label{sec:mbrl_with_joint_loss}
Our approach (Algorithm \ref{algo:mbrl_loop_coupled}) alternates between model learning and data collection. $g$ and $f$ are randomly initialized at the beginning of the learning loop. Each iteration collects data using the controller $g$ for the duration of a predefined horizon $T$. After the roll-out, the collected data is used to update both the forward model $f$ and the controller $g$.
\begin{algorithm}[H]
\small{
\begin{algorithmic}[1]
\STATE{$\mathcal{D} \gets \text{motor babbling data}(s_t, u_t, s_{t+1})$}
\STATE{$f_{\theta} \gets \text{initialize forward model}$}
\STATE{$g_{\beta} \gets \text{initialize inverse model}$}
\STATE{$ \text{train model } f_{\theta}  \text{ on } \mathcal{D}$}
\STATE{$ \text{train model } g_{\beta} \text{ on } \mathcal{D}$}
\WHILE{$i < \text{iter}$}
\STATE{$D_\text{new} \gets \text{rollout $g_{\theta}$ on system} (s_t, u_t, s_{t+1})$}
\STATE{$\mathcal{D} =\mathcal{D} \cup D_\text{new}$}
\STATE{$\text{train model } f_{\theta}  \text{ on } \mathcal{D} \text{ with Loss from \eqref{eq:sup_loss}}$}
\STATE{$\text{train model } g_{\beta} \text{ on } \mathcal{D} \text{ with Loss from \eqref{eq:task_loss} or \eqref{eq:joint_loss} }$}
\ENDWHILE
\end{algorithmic}
\caption{Learning control with Coupled Models}
\label{algo:mbrl_loop_coupled}
}
\end{algorithm}
To update the forward model, we use a regular supervised learning objective representing the model prediction error 
\begin{equation}\label{eq:sup_loss}
   \mathcal{L_{\text{\textit{sup}}}}(\theta) = (f_\theta(s_{t},\tau_t) - s_{t+1})^2
\end{equation}
where $s_{t+1}$ is the next state observed on the robot and $f_\theta(s_{t},\tau_t)$ is the next state predicted by $f$.

To learn $g_\beta$, we propose a loss function \jointloss that trades-off actual robot behavior and control performance prediction using the forward model.
We compare it with two other, simpler, approaches: one, \taskloss that only improves control performance prediction using the forward model and a supervised approach that does not use the forward model.

\subsubsection{Comparison - updating $g$ with \taskloss}
The \taskloss computes a learning objective by comparing the prediction of the forward model (that was coupled with the output of $g_\beta$): $s_{t+1}{_{\theta,\beta}} = f_\theta(s_t,g_{\beta}(s_t,s^*_{t+1}))$ with the desired next state $s^*_{t+1}$ 
\begin{equation}\label{eq:task_loss}
    \mathcal{L}_{\text{\taskloss}}(\beta) = (f_\theta(s_t,g_{\beta}(s_t,s^*_{t+1})) - s^*_{t+1})^2 
\end{equation}
This loss evaluates how well the action of $g$ will be able to achieve the desired state $s^*_{t+1}$ by using $f$ to predict the next state. Intuitively, this will lead to the desired behaviour only if the prediction of the forward model is accurate enough, making the learned controller susceptible to model-bias and inaccuracies.

\subsubsection{Comparison - updating $g$ with supervised loss}
Alternatively, a general supervised learning loss can be used, of the form 
\begin{equation}\label{eq:sup_loss_action}
    \mathcal{L_\text{\textit{inverse sup}}}(\beta) = (g_\beta(s_t,s_{t+1}) - \tau^{run}_{t})^2
\end{equation}
where $s_{t+1}$ is the observed next state when executing $\tau^{run}_{t}$ on the robot, and $\tau^{run}_{t}$ is the output of $g_\beta(s_t,s^*_{t+1})$. This loss is the most common in the literature, especially for inverse dynamics learning \cite{camoriano2016incremental,Pathak2017}.
$\mathcal{L_\text{\textit{inverse sup}}}(\beta)$ uses the observed data to update the controller. In contrast to the \taskloss and also our \jointloss, this loss is not goal oriented, but purely tries to learn the state-control relationship by fitting observed data. 

\subsubsection{Updating $g$ with \jointloss}
Our proposed \jointloss accounts for the quality of the dynamics model, by adding a term that compares the predicted next state with the actual next state. \begin{equation}\label{eq:joint_loss}
\begin{split}
   \mathcal{L}_{\text{\jointloss}}(\beta) = (f_\theta(s_t,g_{\beta}(s_t,s^*_{t+1})) - s^*_{t+1})^2 \\+ (f_\theta(s_t,g_{\beta}(s_t,s^*_{t+1})) - s_{t+1})^2    
\end{split}
\end{equation}
where $s_{t+1}$ is the next state observed on the robot.
The \jointloss thus evaluates not only how well $\tau_\beta$ was able to achieve the desired next state (as predicted by the forward model), but also how good the predictive performance of the forward model actually is. This essentially creates a trade-off between controller and forward model performance, shifting the data distribution seen during roll-out towards a solution that is desirable in reality. In all cases, the parameters of $g_{\beta}$ are then optimized with gradient descent by taking the gradient $\nabla_\beta \mathcal{L}(\beta) $.

In the next section, we analyse in details the \taskloss and the \jointloss. We show why adding the forward model prediction error benefits the controller, and as a consequence, also forward model learning. We then experimentally compare in Section \ref{sec:experiments} these losses with the supervised loss, and show the benefits of our \jointloss.

\begin{figure*}[h]
\centering
\begin{subfigure}[b]{0.32\textwidth}
    \centering
    \includegraphics[width=1.0\textwidth]{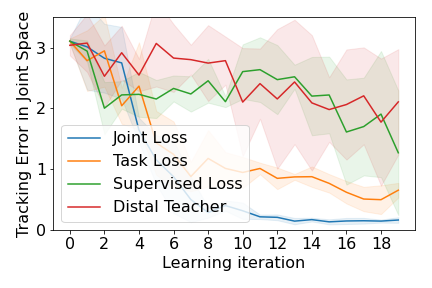}
    \caption{Inverse dynamics learning}
    \label{fig:exp_inverse_dynamics_kuka}
\end{subfigure}
\begin{subfigure}[b]{0.32\textwidth}
    \centering
    \includegraphics[width=1.0\textwidth]{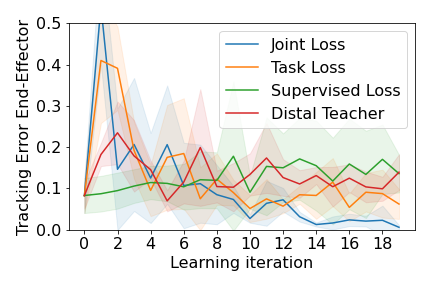}
    \caption{operational-space controller learning}
    \label{fig:exp_op_space_control_kuka}
\end{subfigure}
\begin{subfigure}[b]{0.32\textwidth}
    \centering
    \includegraphics[width=1.0\textwidth]{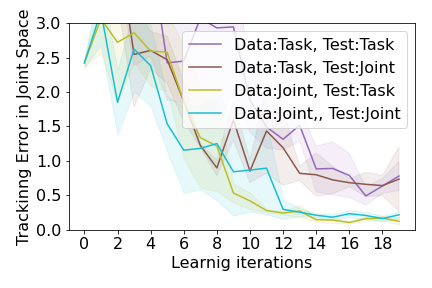}
    \caption{Comparing losses on collected datasets}
    \label{fig:comparison learning from data}
\end{subfigure} 
\caption{Experiments on 7 DoF Kuka arm, the MSE tracking errors (mean and standard deviation over all learning experiments) are reported over learning iterations.}
\label{fig:my_label}
\vspace{-12pt}
\end{figure*}
\subsection{Theoretical analysis of loss functions}
To show the benefit of including the forward model prediction error during inverse model learning let's consider a simplified 1D example: $s_p = f_\theta(s,g_\beta(s,s_d))$.
Where $s_p$ is the prediction of the forward model $f_\theta$, $s$ is the current state, $s_a$ is the actual next state observed on the robot and $s_d$ is the desired next state. $g_\beta$ computes the action for given $s$ and $s_d$.
In order to update parameters $\beta$ of $g$ the gradients that have to be computed are
\begin{equation}\label{eq:task_loss_gradient}
\nabla_\beta \mathcal{L}_{\text{\taskloss}} = 2\frac{\delta f_\theta}{\delta g_\beta}\frac{\delta g_\beta}{\delta \beta}(s_p - s_d)
\end{equation}
and
\begin{equation}\label{eq:joint_loss_gradient}
\nabla_\beta \mathcal{L}_{\text{\jointloss}} = 2\frac{\delta f_\theta}{\delta g_\beta}\frac{\delta g_\beta}{\delta \beta}(2s_p - s_d - s_a)    
\end{equation}
When looking at \eqref{eq:task_loss_gradient} it becomes clear, that $\nabla_\beta \mathcal{L}_{\text{\taskloss}}=0$ when $s_d=s_p$ which means, when the predicted next state is equal to the desired next state. This is a desirable equilibrium, if the forward model prediction is accurate enough, meaning that the predictions of $f$ are not biased. However, if this is not the case, $g$ reaches its equilibrium given a biased model and converges to the wrong solution. We are going to show in section \ref{sec:experiments} how this model bias can affect negatively the learning performance, even if the forward model keeps being improved.

In the case of \eqref{eq:joint_loss_gradient}, the general solution for equilibrium is $s_p=\frac{s_d+s_a}{2}$, which is the average between the desired next state and the measured next state. The special solution $s_p=s_d=s_a$ would be desired. However, since we optimize in an iterative way, if $s_p=\frac{s_d+s_a}{2}$ and we continue optimizing, we can plug the general solution back into $\mathcal{L}_\text{\jointloss}$ and we get 
\begin{equation}
\begin{split}
  \mathcal{L}_\text{\jointloss} = (\frac{s_a+s_d-2s_d}{2})_{\beta,\theta}^2 \\+ (\frac{s_a+s_d-2s_a}{2})_{\beta,\theta}^2 = \frac{1}{2}(s_a-s_d)_{\beta,\theta}^2      
\end{split}
\end{equation}
This means, the loss will reach its global minimum when $s_d=s_a$ which becomes an unbiased loss function. It is worthwhile noting that this loss still carries gradient information for $\beta$ to further improve the inverse model, and is directly affected, through the forward model, by changes of the inverse model. The general solution, $s_p=\frac{s_d+s_a}{2}$, is a local minimum, that the optimization could get stuck in. However we observe that, because our approach alternates between learning the models and collecting new data, the \jointloss and its trade-off between controller performance and forward model prediction error, facilitates data collection that allows to reach the global minimum $s_d=s_a$. We show empirical evidence for this hypothesis in section \ref{sec:experiments}. In addition to being an unbiased loss, this loss also now reflects a kind of feedback controller loss, trying to push the inverse model to match the observed data with the desired data. Once $s_a=s_d$ then also the special solution $s_p=s_d=s_a$ holds and in particular also $s_p = s_a$.
\subsection{Comparison distal teacher loss proposed by \cite{jordan1992forward}}\label{sec:loss_comp_distal}
In \cite{jordan1992forward} the authors propose a stochastic gradient of the form 
\begin{equation}
 \nabla_\beta \mathcal{L}_{\text{\cite{jordan1992forward}}} = \frac{\delta f_\theta}{\delta g_\beta}\frac{\delta g_\beta}{\delta \beta}(s_d - s_a)   
\end{equation}
Here the current gradient of the forward model w.r.t. $\beta$ is used, but the loss does not carry gradient information, as it is purely specified in terms of the observed and desired data. This is equivalent of formulating a loss of the form 
\begin{equation}
  \mathcal{L}_{\text{\cite{jordan1992forward}}} = (s_p - s_d)^2 - (s_p - s_a)^2
\end{equation}
which effectively subtracts the forward model prediction error from the \taskloss error and eventually does not care about the quality of the forward model, as long as the loss between actual and desired next state is decreasing. In simpler scenarios this can have the effect that goal oriented behaviour is achieved even if the forward model is not perfect \cite{jordan1992forward}. On the other hand, this loss does not account for wrong gradients taken through the forward model, that way biasing the solution because of an inaccurate forward model. In practice this seems to be a significant drawback for higher dimensional systems as we show in Sec. \ref{sec:experiments}.

\section{Experiments}\label{sec:experiments}
In this section, we present experiments to show empirically the benefits of learning control with coupled models and our \jointloss. We show how our method of including the forward model prediction error during controller learning outperforms all the other methods. We show evidence that including the forward model prediction error leads to a robot behaviour that favours data collection to improve forward model learning and ultimately less biased models. We perform experiments in simulation with a 7DoF iiwa Kuka arm \cite{kuka} and  the 12 DoF quadruped robot Solo \cite{grimminger2020open} (Fig.\ref{fig:coupled_models_overview}).
All robots are simulated with PyBullet \cite{pybullet}.
\begin{figure*}[h]
    \centering
    \begin{subfigure}[b]{0.23\textwidth}
    \includegraphics[width=1.0\textwidth]{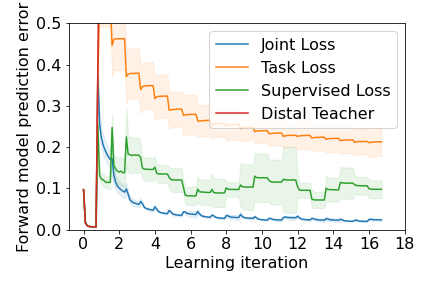}
    \caption{Forward model prediction error during learning(walking)}
    \label{fig:solo_f_model_pred_error}
    \end{subfigure}
    \begin{subfigure}[b]{0.23\textwidth}
    \includegraphics[width=1.0\textwidth]{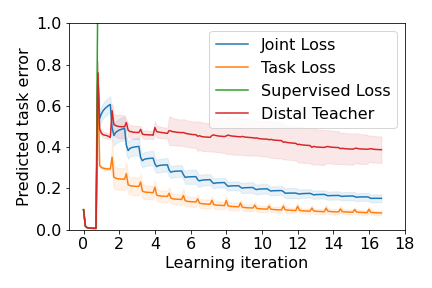}
    \caption{Predicted task error during learning(walking)}
    \label{fig:solo_pred_task_error}
    \end{subfigure}
    \begin{subfigure}[b]{0.23\textwidth}
    \includegraphics[width=1.0\textwidth]{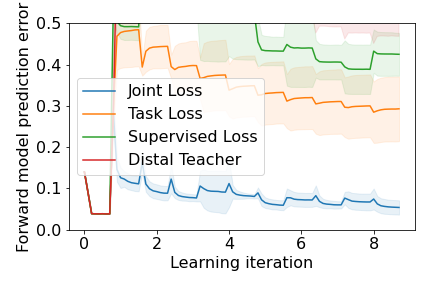}
    \caption{Forward model prediction error during learning(Jump)}
    \label{fig:solo_jump_f_model_pred_error}
    \end{subfigure}
    \begin{subfigure}[b]{0.23\textwidth}
    \includegraphics[width=1.0\textwidth]{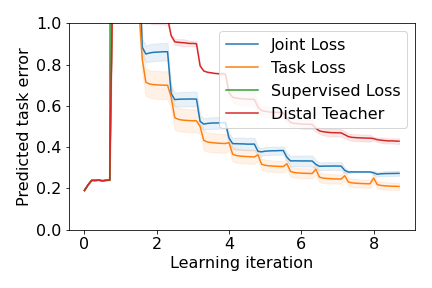}
    \caption{Predicted task error during learning(Jump)}
    \label{fig:solo_jump_pred_task_error}
    \end{subfigure}
\caption{Comparing forward model prediction error and predicted task error during inverse model optimization for walking and jumping} 
\label{fig:solo_inv_model_opt_pred_task}
\vspace{-12pt}
\end{figure*}
\subsection{Experiments with Kuka iiwa}
In these experiments, we learn the forward model with an ensemble of probabilistic neural networks, similar to \cite{chua2018deep}. For the forward model we use three hidden layers with 400 neurons each and ReLU activation functions and an ensemble size of 3. The controller is a neural network with three hidden layers, with 300, 200 and 100 neurons each and ReLU activation function. 

In all these experiments, we compare performance for reaching tasks for five different target positions. Each time, the controller needs to track a desired reaching trajectory (either in joint space or in end-effector space) that is computed in advance, i.e. we learn a tracking controller.

\subsubsection{Inverse dynamics learning}
In this experiments we train the inverse dynamics model of the Kuka arm. The inverse dynamics model $\tau_t=g_\beta(x_t,\ddot{q}^*_{t+1})$ takes as an input the state $s_t=[q_t,\dot{q}_t]$, where $q_t$ are the joint angles and $\dot{q}_t$ the joint velocities at time t, and the desired joint acceleration at the next time step $\ddot{q}^*_{t+1}$ and outputs the torque $\tau_t$. The forward model takes as an input the current state $s_t$ and action $\tau_t$ coming from the inverse model. In Fig.\ref{fig:exp_inverse_dynamics_kuka} we can see that training the inverse model with the \jointloss leads to faster and more stable convergence when compared to the other methods. Notably, it can consistently learn a very good tracking controller in less than 10 iterations.
\subsubsection{Operational space controller learning with forward model as a disambiguator}
In this set of experiments we show how the coupled learning with \jointloss can also be used to learn a task-space controller \cite{khatib1987unified}. $g_\beta$ takes as an input the current state $s_t=[q_t,\dot{q}_t]$ and a desired acceleration in end-effector space $\ddot{x}^*_{{ee}_{t+1}}$. The forward model learns a combination of forward dynamics and kinematics, it takes as an input $s_t$ and $\tau_t$ and outputs $\ddot{q}_t$ and $\ddot{x}_{{ee}_{t+1}}$. The problem of learning an operational space controller is more challenging than learning an inverse dynamics model, since joint redundancy implies that an infinite number of controllers can lead to $\ddot{x}^*_{{ee}_{t+1}}$. 
The non-uniqueness of a perfect tracking controller renders learning difficult when done in a supervised way from collected data, since the same input to $g$ can have different output values. In this scenario, using the coupled models approach, we can disambiguate the problem since the forward model's mapping is unique. In Fig.\ref{fig:exp_op_space_control_kuka}, we can see experimentally how our approach outperforms others enabling to consistently learn an operational space controller in a few iterations. It becomes evident here that learning an operational space controller from data in a supervised learning fashion does not perform satisfactory because of the redundancy in mapping torques to end-effector accelerations. Previous approaches to learn operational space controllers have been proposed \cite{peters2006learning}, however they only consider learning in the vicinity of a local model, to force the selection of only one of the many redundancy resolution strategy. 

\subsubsection{Is the \jointloss improving data collection?}
In model based learning the controller $g$ significantly influences the data seen during learning, since it is the acting component which enables the robot to move and collect more data. 
Therefore, it is natural to ask whether the performance observed with the \jointloss is only due to the quality of the data collected during learning.
To test this hypothesis, we collect two dataset while running our learning loop with \taskloss and \jointloss for inverse dynamics learning. After collecting the two datasets, we re-train the inverse and forward models from scratch with both losses on the datasets collected. The results of this experiment are shown in Fig.\ref{fig:comparison learning from data}, where we can see that the \taskloss and \jointloss perform similarly when they are deployed on the exact same data. In particular, both losses perform better when trained with the data collected using the controller optimized using the \jointloss. This suggests that the data collected while learning with the \jointloss contains more useful information to accomplish a given motor control task. 
\subsection{Experiments on Solo}
\begin{figure}[h]
    \centering
    \includegraphics[width=0.48
    \textwidth]{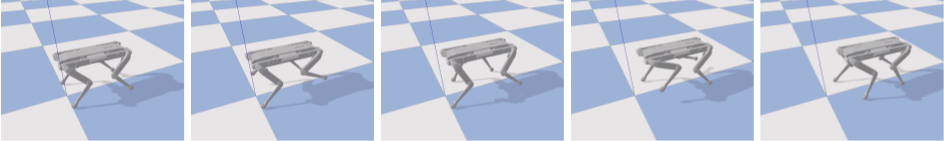}
    \caption{Image sequence of solo successfully walking}
    \label{fig:solo_walking_img_seq}
\end{figure}{} 
In this section we present experiments performed on the Solo quadruped robot \cite{grimminger2020open}. 
The forward model is learned with a neural network with three hidden layers of 1000, 500, 500 neurons each and Relu activation function. The input to the forward model is the current torque $\tau_t$ and the current state $s_t=[x_{{bb}_{t}},q_t,f_{{[x,y,z]}_{t}}, \dot{x}_{{bb}_{t}},\dot{q}_t]$ where $x_{{bb}_{t}}$ is the current base pose (position and orientation), $\dot{x}_{{bb}_{t}}$ the current base velocity and $f_{{[x,y,z]}_{t}}$ are measured contact forces at the four end effectors. The forward model predicts $\Delta s = [\ddot{x}_{{bb}_{t+1}}\Delta t,f_{{[x,y,z]}_{t+1}},\ddot{q}_t\Delta t]$, which is the change in acceleration for the base and joints together with the expected contact forces at $t+1$ when applying $\tau_t$ in $s_t$. The inverse model's architecture is of three hidden layers with 300 neurons each, the input is $s_t$ and $s^*_{t+1}$ which is the desired accelerations of the joints and base together with the desired contact forces at the next time step. We compute the desired trajectories (walking and jumping) using the kino-dynamic planner presented in \cite{ponton_time_2018}. The goal of these experiments is to show that our approach can also use force measurements and handle hard contact switching for unstable underactuated systems.
This is a significantly more challenging task compared to inverse model learning of a fixed base manipulator.
\begin{figure}
\vspace{-3pt}
    \centering
    \begin{subfigure}[b]{0.23\textwidth}
    \centering
    \includegraphics[width=1.0\textwidth]{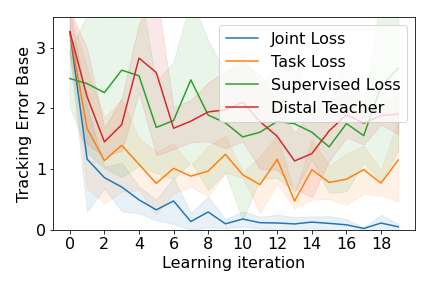}
    \caption{Solo walking}
    \label{fig:solo_walking_base_error}
    \end{subfigure}
    \begin{subfigure}[b]{0.23\textwidth}
    \centering
    \includegraphics[width=1.0\textwidth]{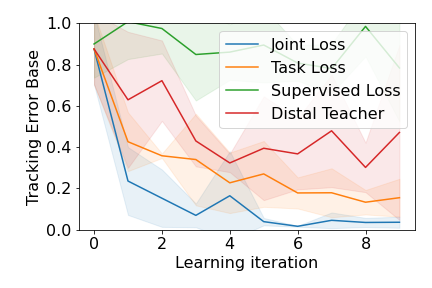}
    \caption{Solo jumping}
    \label{fig:solo_jumping_joint_error}
    \end{subfigure}
    \caption{Inverse model learning on the quadruped, the tracking error in base position and orientation is reported over iterations. Only the \jointloss is successful at the task.}
    \vspace{-17pt}
\end{figure}

\subsubsection{Learning to walk}
In this set of experiments we show how we can learn to control walking. We average our results over 5 different walking horizons of different length and report the mean and the standard deviation of the experiments. When walking, the robot has to make and break contact with the floor multiple times during the trajectory. Making and breaking contact, and transitioning between these two states, is a challenging task that requires careful control at the moment of contact to avoid slipping and preventing the robot from falling. Contact forces are explicitly included in the state as well during model learning and controller optimization.
Thus, the \jointloss also includes the error on the contact forces, between predicted, desired and observed contact forces. 

\begin{figure}[h]
    \centering
    \includegraphics[width=0.45\textwidth]{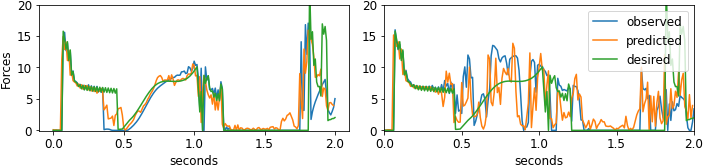}
    \caption{Tracking of ground reaction forces for \jointloss(left) and \taskloss(right) }
    \label{fig:solo_walking_force_error}
\vspace{-5pt}
\end{figure}

In Fig. \ref{fig:solo_walking_base_error} we show the tracking error of the base over learning iterations. We can see clearly here that our \jointloss outperforms all the other approaches. Qualitatively the controller trained with \jointloss is the only one that was able to generate stable walking (cf. Fig.\ref{fig:solo_walking_img_seq}).
This also becomes evident when looking at Table \ref{table:forces}, where we report the tracking error of the ground reaction forces, together with the tracking performance visualized in Fig.\ref{fig:solo_walking_force_error}, where we show the predicted (by the forward model), desired and observed ground reaction forces at the front right foot. The controller trained with the \jointloss (left), tracks the desired forces more accurately than the one trained with \taskloss (right). Also the forward model's prediction of the contact forces, is more accurate in the \jointloss case.
Importantly, the controller is again learned in less than 10 iterations, which makes it amenable to real robot applications.
\vspace{-5pt}
{} 
\subsubsection{Learning to jump}
This experiment shows how Solo can learn to control a jump, 
which is a task with high impact dynamics and requires precise control especially during take-off (to create the right amount of momentum) and landing (to dissipate the impact). We show the results over 5 different jumping heights in Fig.\ref{fig:solo_jumping_joint_error} where we can again see that the \jointloss learns how to successfully accomplish the task. We show again, in Table \ref{table:forces} the tracking error for the ground reaction forces.
\begin{table}
\vspace{3pt}
\tabcolsep=0.11cm
\begin{tabular}{|l|l|l|l|l|}
\hline
                                                                     & \multicolumn{4}{c|}{\textbf{MSE Tracking Error: mean(std) in N}}                                                            \\ \hline
                                                                     & \multicolumn{1}{c|}{\textbf{Joint Loss}} & \textbf{Task Loss} & \textbf{Distal Teacher} & \textbf{Supervised} \\ \hline
\textbf{\begin{tabular}[c]{@{}l@{}}forces \\ (walking)\end{tabular}} & \textbf{0.09 (0.1)}                      & 0.16 (0.18)        & 0.34 (0.35)             & 0.16 (0.26)         \\ \hline
\textbf{\begin{tabular}[c]{@{}l@{}}forces \\ (jumping)\end{tabular}} & \textbf{0.02 (0.01)}                     & 0.3 (0.28)         & 0.38 (0.4)              & 0.28 (4.3)          \\ \hline
\end{tabular}
\caption{Tracking error in ground reaction forces}
\label{table:forces}
\vspace{-17pt}
\end{table}

\subsection{Model bias and its effect on performance}\label{sec:model_bias_illustration}
When looking at Fig. \ref{fig:solo_inv_model_opt_pred_task} we can see how the optimization with \taskloss is prone to find sub-optimal solutions due to a biased forward model. In Fig.\ref{fig:solo_f_model_pred_error} and \ref{fig:solo_jump_f_model_pred_error} the prediction error of the forward model during learning is shown for both experiments. We see that the prediction error of the forward model trained while running the experiment using the \jointloss is lower than the prediction error of the forward model when using the \taskloss. On the other side in Fig.\ref{fig:solo_pred_task_error} and \ref{fig:solo_jump_pred_task_error} the predicted task error ( $ (s_{{t+1}{_{\theta,\beta}}} - s^*_{t+1})^2$ ) is shown. For the experiment trained with the \taskloss, the predicted task error is the lowest, however the prediction error of the forward model is high. This means that the model trained with \taskloss is predicting $s^*_{t+1}$, but the prediction is not correct. This leads to a biased solution, which explains the higher tracking error. Training with the \jointloss does not result in this scenario, since the prediction of the forward model is accurate, in turn, leading to a better performing controller.

\section{CONCLUSIONS}
\vspace{-2pt}
In this work, we show how to leverage forward model prediction error for learning control in an iterative way. Our approach connects controller and forward model learning by using the predicted control signal as an input to the forward model. We show how using forward model prediction error during controller learning results in a learned controller that enables the robot to successfully accomplish non-trivial motor control tasks. We present theoretical and empirical evidence that the improved performance is due to an unbiased loss function, that reduces bias in the learning problem. We also show empirical evidence that when using the controller trained with our \jointloss on the robot, the collected data is more meaningful for the current task and model learning. This could explain the reduced model bias of the forward model during learning control. In simulation, our approach systematically outperforms other approaches also for contact rich tasks on underactuated, unstable systems and enables learning controllers in a few iterations. In future work, we will extend this work for more complex policy learning and test our approach on real robots. 
\bibliography{root}{}

\begin{thebibliography}{10}

\bibitem{kuka}
KUKA AG.
\newblock Kuka ag, 2020.

\bibitem{atkeson1988using}
Christopher~G Atkeson and David~J Reinkensmeyer.
\newblock Using associative content-addressable memories to control robots.
\newblock In {\em Proceedings of the 27th IEEE Conference on Decision and
  Control}, pages 792--797. IEEE, 1988.

\bibitem{Barto2004}
Andrew~G. Barto.
\newblock {Intrinsically motivated learning of hierarchical collections of
  skills}.
\newblock {\em International Conference on Developmental Learning and
  Epigenetic Robotic}, pages 112--119, 2004.

\bibitem{bechtle2020curious}
Sarah Bechtle, Yixin Lin, Akshara Rai, Ludovic Righetti, and Franziska Meier.
\newblock Curious ilqr: Resolving uncertainty in model-based rl.
\newblock In {\em Conference on Robot Learning}, pages 162--171, 2020.

\bibitem{Bellemare2016}
Marc Bellemare, Sriram Srinivasan, Georg Ostrovski, Tom Schaul, David Saxton,
  and Remi Munos.
\newblock Unifying count-based exploration and intrinsic motivation.
\newblock In {\em Advances in Neural Information Processing Systems}, pages
  1471--1479, 2016.

\bibitem{bristow2006survey}
Douglas~A Bristow, Marina Tharayil, and Andrew~G Alleyne.
\newblock A survey of iterative learning control.
\newblock {\em IEEE control systems magazine}, 26(3):96--114, 2006.

\bibitem{camoriano2016incremental}
Raffaello Camoriano, Silvio Traversaro, Lorenzo Rosasco, Giorgio Metta, and
  Francesco Nori.
\newblock Incremental semiparametric inverse dynamics learning.
\newblock In {\em 2016 IEEE International Conference on Robotics and Automation
  (ICRA)}, pages 544--550. IEEE, 2016.

\bibitem{chua2018deep}
Kurtland Chua, Roberto Calandra, Rowan McAllister, and Sergey Levine.
\newblock Deep reinforcement learning in a handful of trials using
  probabilistic dynamics models.
\newblock In {\em NeurIPS}, 2018.

\bibitem{deisenroth2011pilco}
Marc Deisenroth and Carl~E Rasmussen.
\newblock Pilco: A model-based and data-efficient approach to policy search.
\newblock In {\em Proceedings of the 28th International Conference on machine
  learning (ICML-11)}, 2011.

\bibitem{pybullet}
{Erwin Coumans and Yunfei Bai}.
\newblock Pybullet, a python module for physics simulation in robotics, games
  and machine learning.
\newblock \url{http://pybullet.org/}, 2016--2019.

\bibitem{grimminger2020open}
F.~{Grimminger}, A.~{Meduri}, M.~{Khadiv}, J.~{Viereck}, M.~{Wüthrich},
  M.~{Naveau}, V.~{Berenz}, S.~{Heim}, F.~{Widmaier}, T.~{Flayols}, J.~{Fiene},
  A.~{Badri-Spröwitz}, and L.~{Righetti}.
\newblock An open torque-controlled modular robot architecture for legged
  locomotion research.
\newblock {\em IEEE Robotics and Automation Letters}, 5(2):3650--3657, 2020.

\bibitem{herzog2016momentum}
Alexander Herzog, Nicholas Rotella, Sean Mason, Felix Grimminger, Stefan
  Schaal, and Ludovic Righetti.
\newblock Momentum control with hierarchical inverse dynamics on a
  torque-controlled humanoid.
\newblock {\em Autonomous Robots}, 40(3):473--491, 2016.

\bibitem{ishikawa2016cerebro}
Takahiro Ishikawa, Saeka Tomatsu, Jun Izawa, and Shinji Kakei.
\newblock The cerebro-cerebellum: Could it be loci of forward models?
\newblock {\em Neuroscience research}, 104:72--79, 2016.

\bibitem{ito1970neurophysiological}
Masao Ito.
\newblock Neurophysiological aspects of the cerebellar motor control system.
\newblock {\em Int. J. Neurol.}, 7:126--179, 1970.

\bibitem{jarquin_task_hierarchy2013}
G.~{Jarquín}, A.~{Escande}, G.~{Arechavaleta}, T.~{Moulard}, E.~{Yoshida}, and
  V.~{Parra-Vega}.
\newblock Real-time smooth task transitions for hierarchical inverse
  kinematics.
\newblock In {\em 2013 13th IEEE-RAS International Conference on Humanoid
  Robots (Humanoids)}, pages 528--533, 2013.

\bibitem{jordan1992forward}
Michael~I Jordan and David~E Rumelhart.
\newblock Forward models: Supervised learning with a distal teacher.
\newblock {\em Cognitive science}, 16(3):307--354, 1992.

\bibitem{khatib1987unified}
Oussama Khatib.
\newblock A unified approach for motion and force control of robot
  manipulators: The operational space formulation.
\newblock {\em IEEE Journal on Robotics and Automation}, 3(1):43--53, 1987.

\bibitem{koert2018learning}
Dorothea Koert, Guilherme Maeda, Gerhard Neumann, and Jan Pcters.
\newblock Learning coupled forward-inverse models with combined prediction
  errors.
\newblock In {\em 2018 IEEE International Conference on Robotics and Automation
  (ICRA)}, pages 2433--2439. IEEE, 2018.

\bibitem{laversanne2018curiosity}
Adrien Laversanne-Finot, Alexandre Pierre, and Pierre-Yves Oudeyer.
\newblock Curiosity driven exploration of learned disentangled goal spaces.
\newblock {\em arXiv preprint arXiv:1807.01521}, 2018.

\bibitem{lee2019making}
M.~A. {Lee}, Y.~{Zhu}, K.~{Srinivasan}, P.~{Shah}, S.~{Savarese}, L.~{Fei-Fei},
  A.~{Garg}, and J.~{Bohg}.
\newblock Making sense of vision and touch: Self-supervised learning of
  multimodal representations for contact-rich tasks.
\newblock In {\em 2019 International Conference on Robotics and Automation
  (ICRA)}, pages 8943--8950, 2019.

\bibitem{levine2013guided}
Sergey Levine and Vladlen Koltun.
\newblock Guided policy search.
\newblock In {\em International Conference on Machine Learning}, pages 1--9,
  2013.

\bibitem{lopes2012exploration}
Manuel Lopes, Tobias Lang, Marc Toussaint, and Pierre-Yves Oudeyer.
\newblock Exploration in model-based reinforcement learning by empirically
  estimating learning progress.
\newblock In {\em Advances in neural information processing systems}, pages
  206--214, 2012.

\bibitem{lutter2019deep}
Michael Lutter, Christian Ritter, and Jan Peters.
\newblock Deep lagrangian networks: Using physics as model prior for deep
  learning.
\newblock {\em arXiv preprint arXiv:1907.04490}, 2019.

\bibitem{miall1996forward}
R~Chris Miall and Daniel~M Wolpert.
\newblock Forward models for physiological motor control.
\newblock {\em Neural networks}, 9(8):1265--1279, 1996.

\bibitem{miller1987sensor}
W~Miller.
\newblock Sensor-based control of robotic manipulators using a general learning
  algorithm.
\newblock {\em IEEE Journal on Robotics and Automation}, 3(2):157--165, 1987.

\bibitem{nguyen2008computed}
Duy Nguyen-Tuong, Matthias Seeger, and Jan Peters.
\newblock Computed torque control with nonparametric regression models.
\newblock In {\em 2008 American Control Conference}, pages 212--217. IEEE,
  2008.

\bibitem{Pathak2017}
Deepak Pathak, Pulkit Agrawal, Alexei~A. Efros, and Trevor Darrell.
\newblock {Curiosity-Driven Exploration by Self-Supervised Prediction}.
\newblock {\em IEEE Computer Society Conference on Computer Vision and Pattern
  Recognition Workshops}, 2017-July:488--489, 2017.

\bibitem{peters2006learning}
Jan Peters and Stefan Schaal.
\newblock Learning operational space control.
\newblock In {\em Robotics: Science and Systems}, 2006.

\bibitem{ponton_time_2018}
Brahayam Ponton, Alexander Herzog, Andrea Del~Prete, Stefan Schaal, and Ludovic
  Righetti.
\newblock On time optimization of centroidal momentum dynamics.
\newblock In {\em 2018 {IEEE} {International} {Conference} on {Robotics} and
  {Automation} ({ICRA})}, pages 5776--5782, Brisbane, Australia, 2018. IEEE.

\bibitem{schillaci2012coupled}
Guido Schillaci, Verena~V Hafner, and Bruno Lara.
\newblock Coupled inverse-forward models for action execution leading to
  tool-use in a humanoid robot.
\newblock In {\em 2012 7th ACM/IEEE International Conference on Human-Robot
  Interaction (HRI)}, pages 231--232. IEEE, 2012.

\bibitem{shyammax}
Pranav Shyam, Wojciech Jaskowski, and Faustino Gomez.
\newblock Model-based active exploration.
\newblock {\em arXiv preprint arXiv:1810.12162}, 2018.

\bibitem{Singh2004}
S.~Singh, A.G. Barto, and N.~Chentanez.
\newblock {Intrinsically motivated reinforcement learning}.
\newblock {\em 18th Annual Conference on Neural Information Processing Systems
  (NIPS)}, 2004.

\bibitem{Singh2010}
Satinder Singh, Richard~L. Lewis, Andrew~G. Barto, and Jonathan Sorg.
\newblock {Intrinsically Motivated Reinforcement Learning: An Evolutionary
  Perspective}.
\newblock {\em IEEE Transactions on Autonomous Mental Development},
  2(2):70--82, 2010.

\bibitem{talvitie2014model}
Erik Talvitie.
\newblock Model regularization for stable sample rollouts.
\newblock In {\em UAI}, pages 780--789, 2014.

\bibitem{talvitie2016self}
Erik Talvitie.
\newblock Self-correcting models for model-based reinforcement learning.
\newblock {\em arXiv preprint arXiv:1612.06018}, 2016.

\bibitem{tanneberg2019intrinsic}
Daniel Tanneberg, Jan Peters, and Elmar Rueckert.
\newblock Intrinsic motivation and mental replay enable efficient online
  adaptation in stochastic recurrent networks.
\newblock {\em Neural Networks}, 109:67--80, 2019.

\bibitem{wolpert1995internal}
Daniel~M Wolpert, Zoubin Ghahramani, and Michael~I Jordan.
\newblock An internal model for sensorimotor integration.
\newblock {\em Science}, 269(5232):1880--1882, 1995.

\bibitem{wolpert1998multiple}
Daniel~M Wolpert and Mitsuo Kawato.
\newblock Multiple paired forward and inverse models for motor control.
\newblock {\em Neural networks}, 11(7-8):1317--1329, 1998.

\bibitem{wolpert1998internal}
Daniel~M Wolpert, R~Chris Miall, and Mitsuo Kawato.
\newblock Internal models in the cerebellum.
\newblock {\em Trends in cognitive sciences}, 2(9):338--347, 1998.

\bibitem{zhang2019leveraging}
Kevin Zhang, Mohit Sharma, Manuela Veloso, and Oliver Kroemer.
\newblock Leveraging multimodal haptic sensory data for robust cutting.
\newblock In {\em Proceedings of IEEE-RAS International Conference on Humanoid
  Robots}, October 2019.

\end{thebibliography}
\bibliographystyle{plain}

\end{document}